\documentclass[sn-mathphys-num,iicol]{sn-jnl}
\usepackage{graphicx}%
\usepackage{multirow}%
\usepackage{amsmath,amssymb,amsfonts}%
\usepackage{amsthm}%
\usepackage{mathrsfs}%
\usepackage[title]{appendix}%
\usepackage{xcolor}%
\usepackage{textcomp}%
\usepackage{manyfoot}%
\usepackage{booktabs}%
\usepackage{algorithm}%
\usepackage{algorithmicx}%
\usepackage{algpseudocode}%
\usepackage{listings}%
\theoremstyle{thmstyleone}%
\theoremstyle{thmstyletwo}%

\theoremstyle{thmstylethree}%

\begin{document}

\title[Article Title]{An Analysis for Image-to-Image Translation and Style Transfer}


\author[1,2]{\fnm{Xiaoming} \sur{Yu}}\email{yuxiaoming2023@ia.ac.cn}

\author[1,2]{\fnm{Jie} \sur{Tian}}

\author[1,2]{\fnm{Zhenhua} \sur{Hu}}

\affil[1]{\orgdiv{CAS Key Laboratory of Molecular Imaging, Beijing Key Laboratory of Molecular Imaging, Institute of Automation}, \orgname{Chinese Academy of Sciences}}
\affil[2]{\orgdiv{School of Artificial Intelligence}, \orgname{University of Chinese Academy of Sciences}}


\abstract{With the development of generative technologies in deep learning, a large number of image-to-image translation and style transfer models have emerged at an explosive rate in recent years. These two technologies have made significant progress and can generate realistic images. However, many communities tend to confuse the two, because both generate the desired image based on the input image and both cover the two definitions of content and style. In fact, there are indeed significant differences between the two, and there is currently a lack of clear explanations to distinguish the two technologies, which is not conducive to the advancement of technology. We hope to serve the entire community by introducing the differences and connections between image-to-image translation and style transfer. The entire discussion process involves the concepts, forms, training modes, evaluation processes, and visualization results of the two technologies. Finally, we conclude that image-to-image translation divides images by domain, and the types of images in the domain are limited, and the scope involved is small, but the conversion ability is strong and can achieve strong semantic changes. Style transfer divides image types by single image, and the scope involved is large, but the transfer ability is limited, and it transfers more texture and color of the image.}

\maketitle
%%%%%%%%%%%%%%%%%%%%%%%%%%%%%%%%%%%%%%%%%%%%%%%%%%%%%%%%%%%%%%%%%%%%%%%%%%%%%%%%%%%%%

\section{Introduction}

\begin{figure*}
\centering
\includegraphics[width=0.8\textwidth]{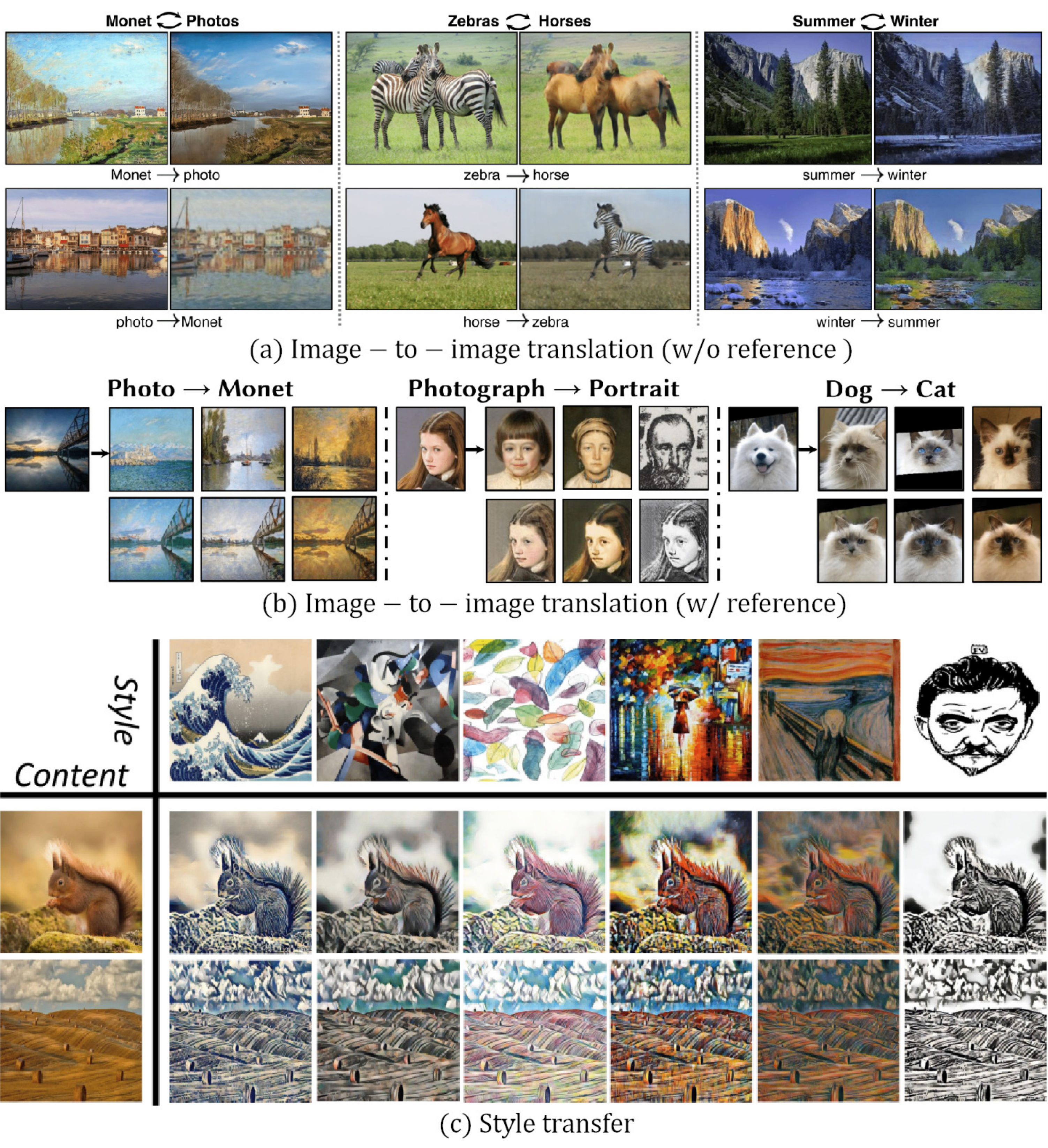} % Reduce the figure size so that it is slightly narrower than the column.
\caption{(a) Image-to-image translation (w/o reference). In each set of images, the left side is the input image and the right side is the generated image. The six tasks shown require a total of six models. The models and generated results are from CycleGAN \cite{zhu2017unpaired}. (b) Image-to-image translation (w/ reference). In each group of images, the first row is the reference image, and the second row is the generated image. The three tasks shown require a total of three models. The models and generated results are from DSMAP \cite{chang2020domain}. (c) Style transfer. The first column is the content image, the first row is the style reference image, and the rest are generated images. The generation task shown requires a total of one model. The model and generation results are from AdaAttN \cite{liu2021adaattn}.}
\label{fig1}
\end{figure*}

In recent years, image generation has become a hot topic and has received increasing attention from researchers. Image-to-image translation \cite{zhu2017unpaired,chang2020domain,liang2021high} and style transfer \cite{liu2021adaattn,huang2017arbitrary,park2019arbitrary} are two of the main branches. After the efforts of a large number of researchers, great results have been achieved, and the generated images are almost indistinguishable from the real ones \cite{zhu2017unpaired,chang2020domain,liu2021adaattn}, as shown in Figure \ref{fig1}. Although with the emergence of diffusion model \cite{ho2020denoising}, generative models have become more and more brilliant, and multi-modal generation technologies have emerged in the community, such as the combination of text and image/video \cite{ramesh2022hierarchical,liu2024sora,hoe2024interactdiffusion}, which makes single-modal image-to-image translation and style transfer slightly inferior, these two technologies, as the research foundation and pioneers, provide conditions for current multi-modal generation, and are still not fully understood. Therefore, we analyze and discuss these two technologies in this paper. Image translation is the conversion of an image from one domain to another. In one case, the model directly converts the input image from the source domain to the target domain \cite{zhu2017unpaired,lin2022exploring}, and in another case, the conversion from the source domain to the target domain is completed with the reference of the target domain image \cite{chang2020domain,huang2018multimodal}. Style transfer only exists in one form, where the model generates a stylized image from the content image with the style image as a reference. The different generation forms of the two technologies are shown in Figure \ref{fig1}. Since both technologies use reference images as conditions to achieve image transformation, many people are confused about the connection and difference between the two technologies, and even confuse the two technologies. Although the forms of the two technologies are similar, they are essentially different. The goal of this paper is to introduce the two technologies in detail to reveal their similarities and differences. However, please note that this paper is designed for researchers with a certain background in this field to help them improve their technology. It cannot be used as an introduction for beginners, because we mainly focus on the differences and connections between the two technologies.

The structure of this paper first includes a background introduction to image-to-image translation and style transfer. Next, the technologies involved in the two are elaborated, and finally their similarities and differences are discussed.

\section{Image-to-Image Translation}

\begin{figure}
\centering
\includegraphics[width=\columnwidth]{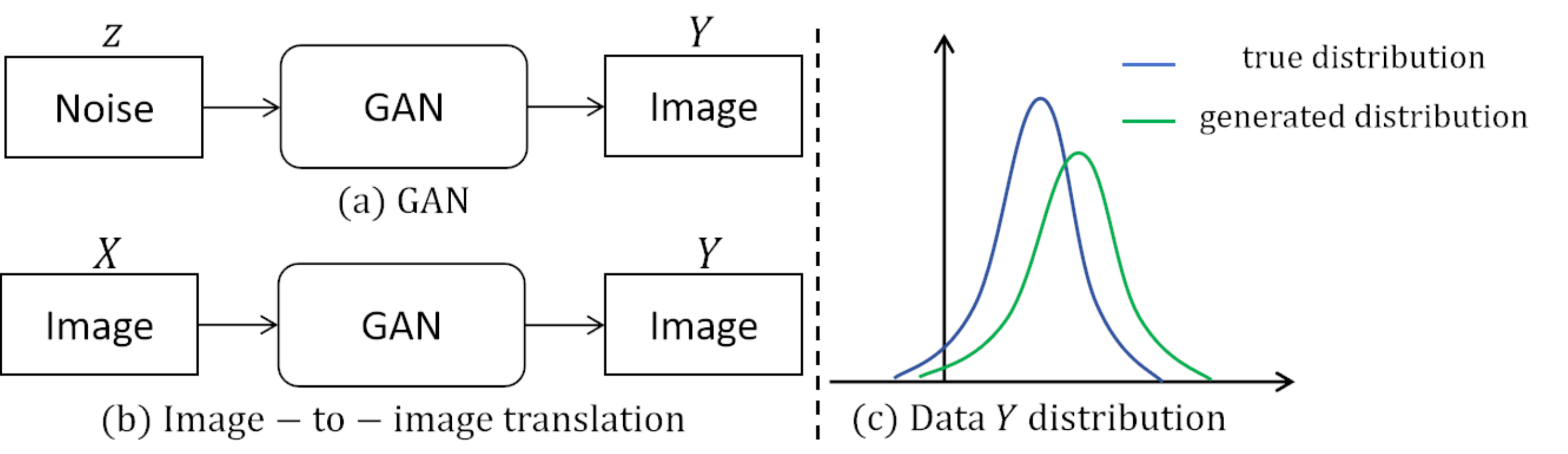} % Reduce the figure size so that it is slightly narrower than the column.
\caption{(a) The generation process of the generative adversarial network. The noise $z$ is converted to the image domain $Y$ through the model. (b) The generation process of image-to-image translation. The image domain $X$ is converted to the image domain $Y$ through the model. (c) Data distribution. The distribution of the generated images and the real images.}
\label{fig2}
\end{figure}

The emergence of generative adversarial network \cite{goodfellow2014generative} has led to the application of image-to-image translation. Prior to this, the application of variational auto-encoder \cite{larsen2016autoencoding} is a prototype, and supervised image translation is Pix2Pix \cite{isola2017image}, which appears in 2017. As can be seen from Figure \ref{fig2}, the original generative adversarial network model generates images from noise, and the goal of the model is to make the distribution of generated images close to the distribution of real images. A similar expression, image-to-image translation, assumes that there are one or more domains $\Omega_i\vert_{i=1,2,...}$. Based on a certain domain $\Omega_x$, an image $\omega_i^{x}$ of the domain is collected as an input image. The generative model can convert the input image into an image $\omega_i^{y}$ of another domain $\Omega_y$, so that the image generated by the model is close to the distribution of the image in the $\Omega_y$ domain.

Image-to-image translation divides images by domain. Images within the entire domain have similar content and style features, and images between domains have similar content features. Common tasks include the transformation of seasons \cite{romero2019smit,yu2019multi}, the transformation of human faces \cite{kim2019u,lee2018diverse}, and the transformation of animal faces \cite{kim2022style,ko2022self}, etc. When the content gap between domains is large, such as the conversion between animal faces and human faces, carefully designed structures need to be added to meet the requirements \cite{liu2021separating}. General basic methods can only achieve the transformation from one domain to another through a single model \cite{liang2021high,kim2019u}, such as the transformation from cat face to dog face. If a domain is added, such as tiger face, the number of models needs to be increased, and an additional model for cat face to tiger face translation needs to be trained. Multi-domain image translation combines input images with label information \cite{choi2018stargan,yu2019multi}, or improves the structure of the network \cite{choi2020stargan,li2021image}, so that a single model can achieve conversion in multiple domains, such as cat face to dog face or tiger face. Although the scope of application is expanded, additional consumption or structure is added. At this time, the domain is upper bounded because the network structure cannot be infinitely large. This is also subject to many constraints, such as the need to collect a large number of labeled image categories. Although image-to-image translation is severely constrained by domains, researchers have tried to expand its application through the concept of style \cite{chang2020domain,huang2018multimodal}, which is mainly reflected in the style differences within the domain. Let's still take animal face conversion as an example. Dog faces involve different breeds of dogs, different colors, face shapes, etc. By defining the style in dog faces, a single cat face image can be converted into different dog face images while maintaining the main content structure of the cat face, such as overall posture, eye opening, etc. We distinguish between the two by single-output model (a single input image produces a single output image) and multi-output model (a single input image can generate multiple output images). This difference can be seen in Figure \ref{fig1} (a) and (b).

\begin{figure*}[t]
\centering
\includegraphics[width=0.8\textwidth]{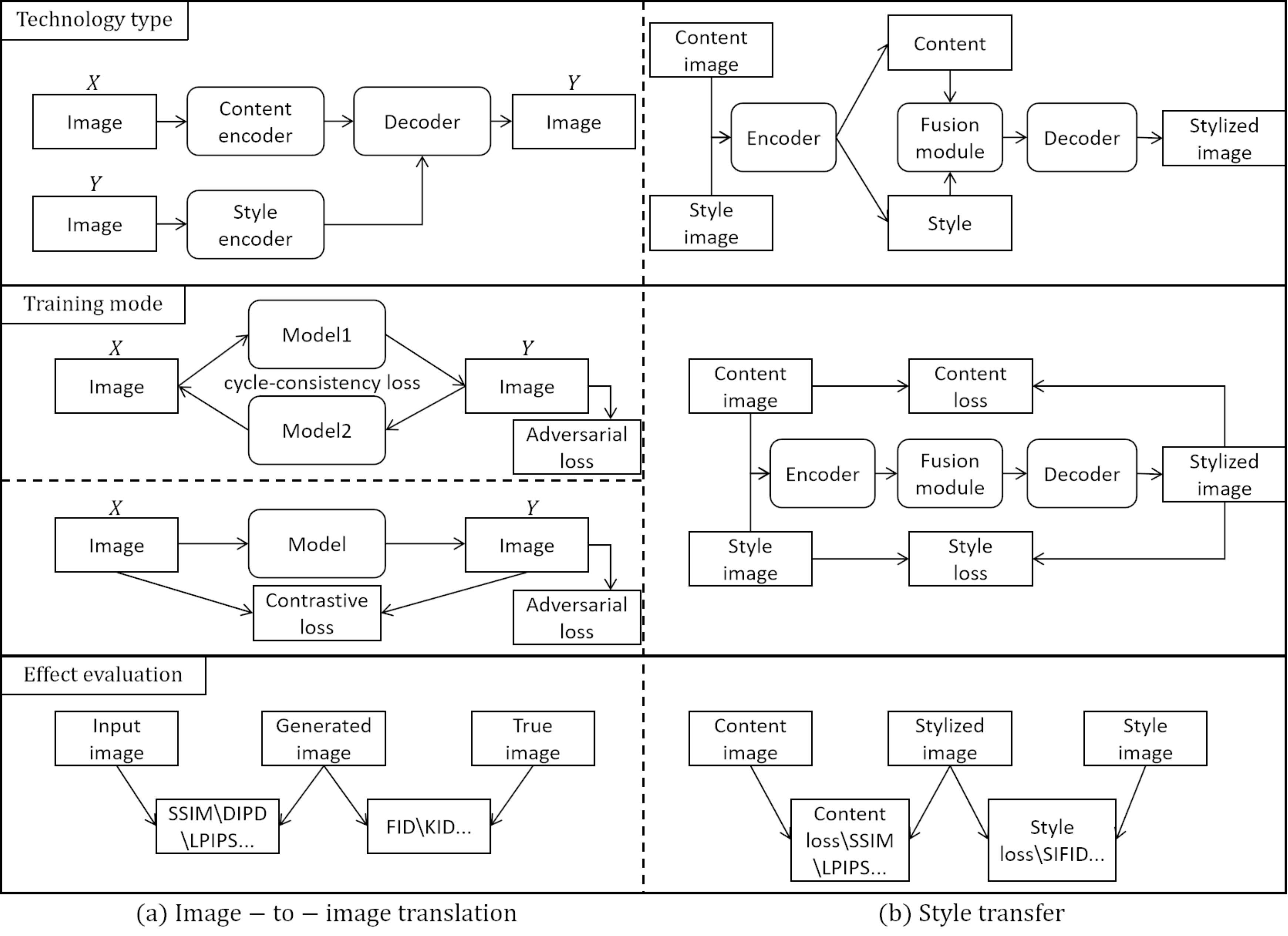} % Reduce the figure size so that it is slightly narrower than the column.
\caption{The comparison between unsupervised image translation and arbitrary style transfer includes three parts: technology type, training mode and effect evaluation. $X$ and $Y$ represent images from two different domains.}
\label{fig3}
\end{figure*}

Image-to-image translation training is mainly carried out in the form of generative adversarial network, because the generated images need to be ensured to be maintained in the target domain. But there is also the earliest form of auto-encoder training \cite{larsen2016autoencoding}. When the data form is paired, the generated images can be constrained by pixel consistency with the real images to constrain the entire process \cite{isola2017image}. But obviously, paired data is limited in nature. At the same time, unsupervised image translation is achieved through cycle consistency loss constraints \cite{zhu2017unpaired}. When the model converts an image from the source domain to the target domain, it is easy to generate meaningless images to confuse the discriminator, and the image structure is destroyed. However, by inverting it back to the source domain through another model and adding pixel consistency loss, the original content features can be constrained not to be lost. Due to the simplicity and convenience of cycle consistency loss, a large number of unsupervised image translation methods \cite{chang2020domain,huang2018multimodal,chen2021mutual,chen2020reusing} have been derived based on CycleGAN \cite{zhu2017unpaired}. In addition, there is no additional model inversion process, but the addition of some losses also maintains the image content structure \cite{benaim2017one,xie2023unpaired}. Another mainstream method to constrain the content to remain unchanged during image translation is through contrastive learning \cite{park2020contrastive}. In this case, there is no need for an additional model to invert the generated image back to the source domain. It is sufficient to maintain the content structure by maintaining the consistency of the input image and the generated image from the perspective of the image patch. Similarly, a large number of methods \cite{hu2022qs,jung2022exploring,zheng2021spatially,wang2021instance} have also been derived from the CUT \cite{park2020contrastive}.

There are two ways to evaluate the quality of images generated by image-to-image translation. One is to use a single image. The degree of preservation of source image structure is an evaluation part in image translation. Content details cannot be lost during the translation process. Different images have different structures, which can only be calculated through a single image. For example, indicators such as the Domain-Invariant Perceptual Distance (DIPD) \cite{huang2018multimodal,liu2019few}, the Structural Similarity Index Measure (SSIM), and the Learned Perceptual Image Patch Similarity (LPIPS) \cite{zhang2018unreasonable} can be used to measure the content similarity between the generated image and the corresponding input image. Another way is to evaluate the quality of the generated image based on the scope of the domain. Image translation divides different types of images by domain. The generated image needs to ensure that it is converted to the target domain, that is, the authenticity of the image needs to be evaluated. Image authenticity evaluation mainly involves two indicators: Frchet Inception Distance (FID) \cite{heusel2017gans} and Kernel Inception Distance (KID) \cite{binkowski2018demystifying}. FID fits the generated images and the real images into a distribution respectively, which is assumed to be a multi-dimensional Gaussian distribution. The authenticity of the generated images can be measured by comparing the mean and covariance of the distribution. KID is similar to FID and is also used to measure the authenticity of a batch of samples.

\section{Style Transfer}

Style transfer is the process of combining a content image and a style image to generate a stylized image, which retains the main structure of the content image and integrates the style representation of the style image. Style transfer is for a single image. Early work focuses on processing a content image and a style image \cite{gatys2016image}, and then on processing arbitrary content images and a style image \cite{dumoulin2016learned,li2017diversified}. Now, it is widely used to rely on a single model to achieve the processing of arbitrary content images and arbitrary style images \cite{huang2017arbitrary,li2019learning,wu2022ccpl}, which is the focus of current research. The data is divided into two parts, the content dataset and the style dataset. Even though there are two parts of data, the technology is for a single image, because the types of images covered in the two datasets are huge and cannot be accurately divided into specific categories. The content dataset includes images of any scene, and the style dataset includes images of any style mode. The commonly used content dataset is MS-COCO \cite{lin2014microsoft}, and the style dataset is Wikiart \cite{phillips2011wiki}, both of which have about 80,000 images. In general, conventional training does not include label information, so it is still a single image processing. This can be seen more clearly from the subsequent training introduction.

\begin{figure*}[t]
\centering
\includegraphics[width=0.75\textwidth]{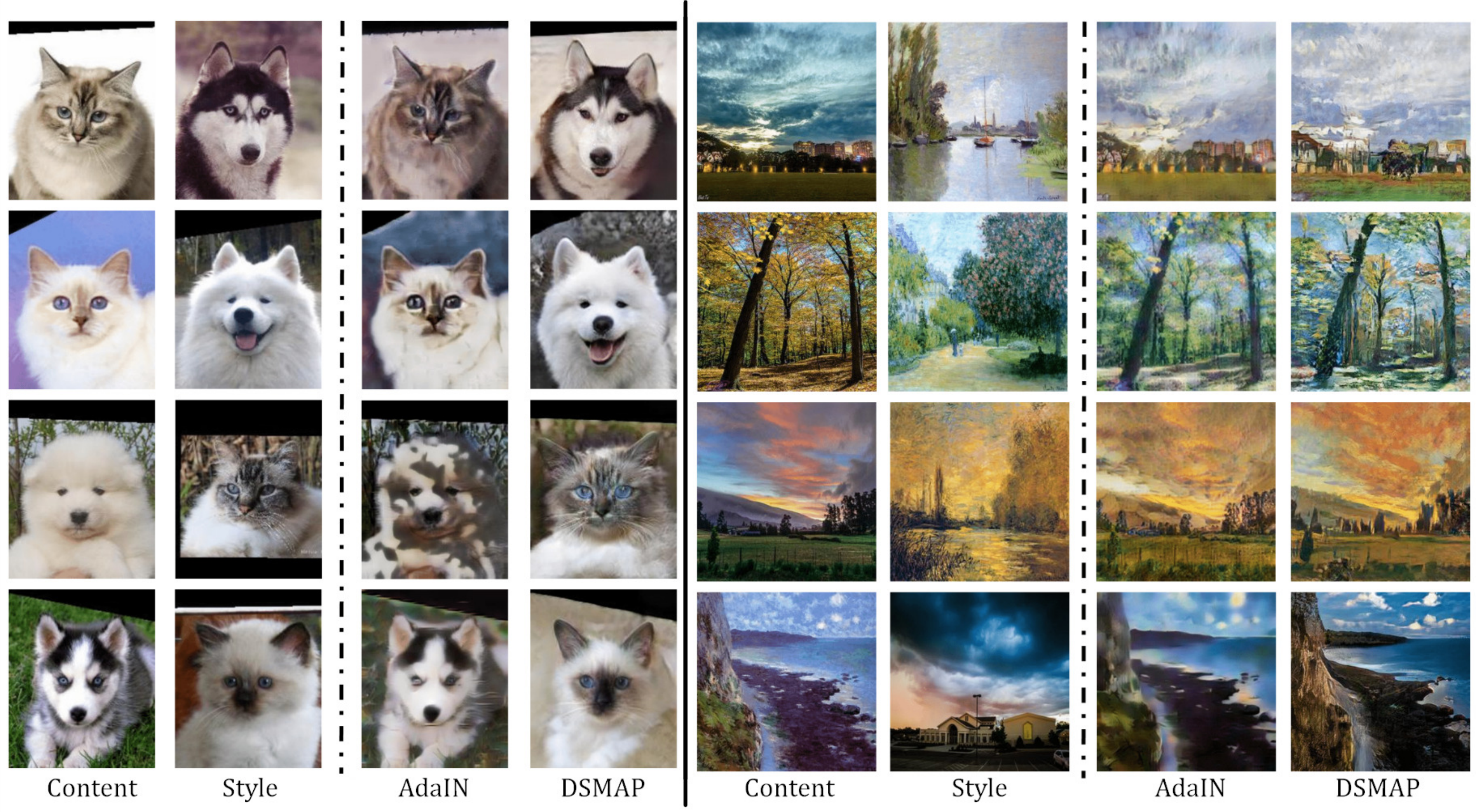} % Reduce the figure size so that it is slightly narrower than the column.
\caption{Comparison of the generative effects of unsupervised image translation and arbitrary style transfer. The left side shows the conversion between cats and dogs, and the right side shows the conversion from photos to Monet paintings. AdaIN \cite{huang2017arbitrary} is a model for arbitrary style transfer, and DSMAP \cite{chang2020domain} is a model for unsupervised image translation. The generative results are derived from DSMAP \cite{chang2020domain}.}
\label{fig4}
\end{figure*}

The processing of a single content image and a single style image is carried out in the form of iterative optimization \cite{gatys2016image}, and the Gram style representation is introduced. The processing of arbitrary content images and single style images is based on feed-forward networks and Gram representation \cite{dumoulin2016learned,li2017diversified}. Arbitrary style transfer often uses content perceptual loss and style perceptual loss as content loss and style loss to process arbitrary content images and arbitrary style images \cite{li2019learning,wu2022ccpl,deng2022stytr2}. Its training mainly exists in the form of auto-encoder, and content and style fusion modules are interspersed between the encoder and decoder. Content feature maps and style feature maps are generated through pre-trained models \cite{simonyan2014very}. Various fusion modules \cite{li2019learning,wu2022ccpl,deng2022stytr2,wu2021styleformer,zheng2024puff} fuse content and style features to obtain stylized feature maps, which are then decoded by a similarly symmetric decoder to obtain stylized images. The content loss and style loss constrain the generated image to be similar in structure to the content image and similar in style to the style image.

The types of content images and style images are complex and it is difficult to clearly define the categories, so there is no unified distribution to measure all images. The evaluation process of style transfer is performed on a single generated image, mainly focusing on two aspects: whether the generated image retains the content structure of the content image well, and whether the generated image integrates the style representation of the style image well. The commonly used content structure evaluation indicators are SSIM from the image level and content loss and LPIPS \cite{zhang2018unreasonable} from the model level. At this time, the similarity between the feature maps of the generated image and the content image is calculated, which measures the consistency of the deep semantics of the image. The commonly used style consistency evaluation indicators are style loss and Single Image FID (SIFID) \cite{shaham2019singan}. Style loss is used to constrain the entire training process and can naturally be used as an evaluation method. Unlike FID, SIFID is redesigned and can be used to evaluate the style consistency of a single generated image and a single style image.

\section{Discussion}

From the introduction of the above two technologies, we can understand that they both use input images as the basis to generate output images. The multi-output image translation model is extremely similar to style transfer. It is this similarity of the process that has caused many groups to confuse the two. However, from the perspective of technology type, training mode and effect evaluation, the two are indeed different. The following is a detailed distinction between the two widely used technologies of unsupervised image translation and arbitrary style transfer. For details, please refer to Figure \ref{fig3} and the following text.

\begin{figure*}
\centering
\includegraphics[width=0.75\textwidth]{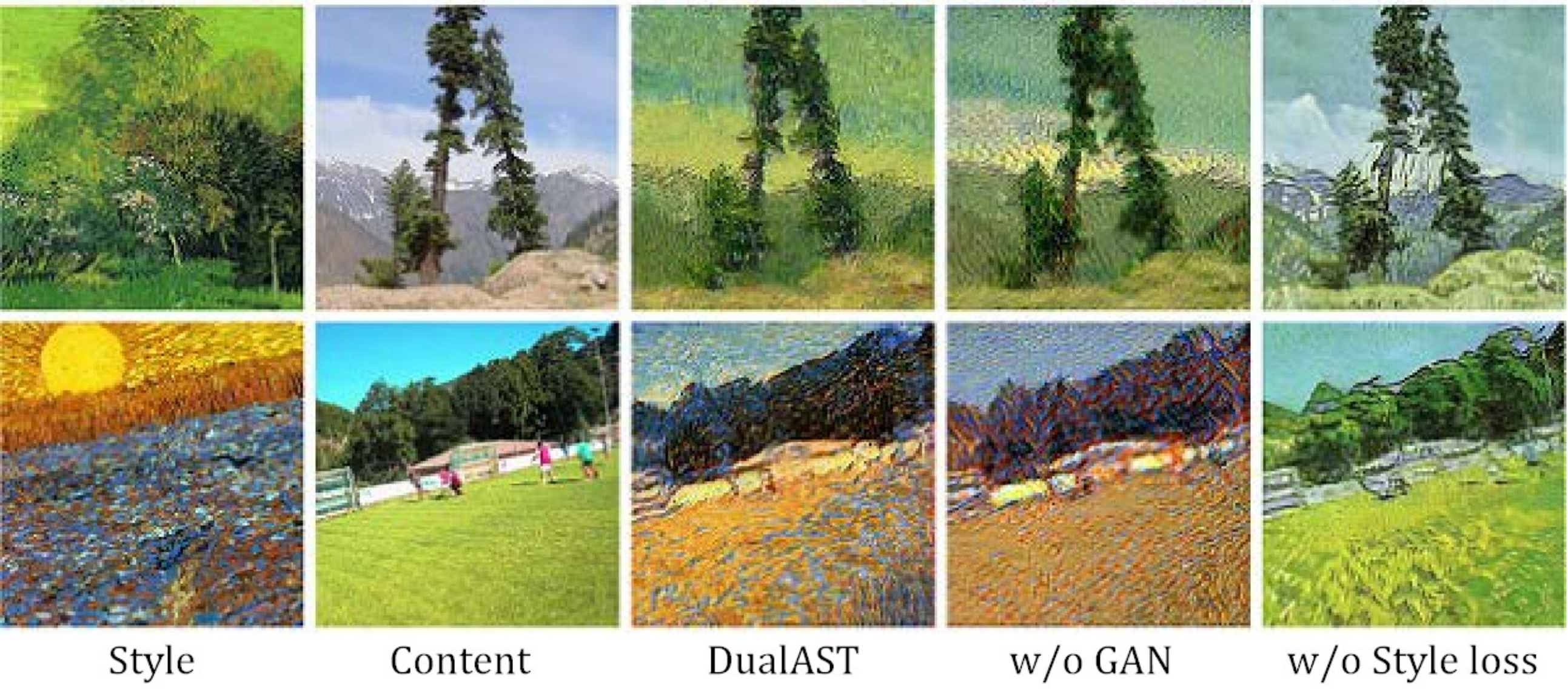} % Reduce the figure size so that it is slightly narrower than the column.
\caption{Comparison of the effects of style loss and adversarial loss. DualAST \cite{chen2021dualast} is a complete model that includes adversarial loss and style loss. w/o GAN retains the style loss and removes the adversarial loss, while w/o Style loss retains the adversarial loss and removes the style loss. The generated results are from DualAST \cite{chen2021dualast}.}
\label{fig5}
\end{figure*}

\textbf{Technology type.} Image translation is to divide images into different categories, corresponding to different domains. Given two domains $X$ and $Y$, the goal is to achieve the conversion from $X$ to $Y$. The images collected in the $X$ domain and the $Y$ domain are input into the content encoder and the style encoder respectively, and the content feature map and the style code are fused in the decoder to generate the image in the $Y$ domain. The multi-output model generates images based on different style in the domain as references. At this time, the content definition is different content of the same category (for example, cats have different body postures and open and closed eyes), and the style definition is different style of the same category (for example, different colors of dog hair and the richness of hair). In arbitrary style transfer, given a content image and a style image, the two images are passed through pre-trained model to obtain the content feature map and the style feature map. These two feature maps are fused in a carefully designed content and style feature map fusion module to obtain a stylized feature map. Finally, the decoder converts the features of the latent space into an image. There are many types of images in style transfer. In a loose sense, each content image is a content structure, which can be any object such as buildings, animals, plants and vehicles. Each style image is a style representation, which can be various colors, styles of various painters and various weather types. There are differences in the content and style definitions in image translation and style transfer. From the perspective of technology type alone, style transfer involves more aspects of style and content, and has a wider range of applications than image translation. The content and style range involved in image translation is smaller but more detailed. Figure \ref{fig4} shows the difference between the two technologies in animal conversion and Monet style painting conversion. Arbitrary style transfer AdaIN \cite{huang2017arbitrary} only shallowly integrates the color and texture of the style part in animal face conversion, and strong semantic related information such as the overall facial shape remains unchanged. Image translation DSMAP \cite{chang2020domain} has shown strong advantages in domain conversion and can achieve semantically related deformations. In the generation of the Monet style painting on the right, due to the lower semantic requirements, the two technologies show little difference and the generated images are very similar. However, DSMAP only trained the model for a single scene, Monet. AdaIN can be trained and generated in a wider range of artistic style paintings, and has weak constraints on categories in the style, reflecting the advantages of style transfer.

\textbf{Training mode.} Supervised image translation ensures that the generated image is consistent with the ground truth through pixel consistency constraints. Direct unsupervised training of the model can easily generate meaningless images. The overall structure of the image is destroyed, but the discriminator cannot distinguish it \cite{zhu2017unpaired}. Unsupervised image translation requires the addition of specific structures to ensure that the generated image is reasonable. The two main modes are to apply cycle consistency loss through an additional reverse conversion model \cite{zhu2017unpaired} or to rely on contrastive loss to constrain the consistency of different patches of the image \cite{park2020contrastive}. The image is converted to the target domain using adversarial loss constraints. The discriminator is trained on both the target domain images and the generated images, and learns the overall style of the target domain images, which has the ability to guide the generator training. Most of the large group of image translation is designed based on generative adversarial networks. Style transfer is different. It only has an auto-encoder structure, in which the encoder is a pre-trained model such as VGG-19 \cite{simonyan2014very}, and the decoder is a trainable structure similar to VGG-19. Most methods seek a fusion method with advantageous content and style in the latent space between the two to obtain a stylized feature map. The optimization of content and style loss is calculated for a single image. The content structure of the stylized image is consistent with the content image through the content loss constraint, and the style texture of the stylized image is consistent with the style image through the style loss constraint. There is no need to consider more modules and methods to constrain the training process. Some other models \cite{chen2021dualast,chen2021artistic} combine style loss and adversarial loss based on arbitrary style transfer in order to utilize the ability of adversarial loss to learn the overall style. As shown in Figure \ref{fig5}, the arbitrary style transfer model is trained in the data sets Place365 \cite{zhou2014learning} and Wikiart \cite{phillips2011wiki}. When only adversarial loss exists (w/o Style loss), due to the wide range of style categories, the generative adversarial network model is difficult to capture the style distribution of an entire domain, and the style texture of the generated image does not conform to the style image. This also shows that the model trained in the form of image translation in a multi-category style dataset cannot accurately reflect multi-category style information, and the application scenarios of image translation are limited. The model trained in the form of style transfer (w/o GAN) can accurately reflect the style of the style image. This shows the wide range of application scenarios in style transfer. Based on the style loss, adding adversarial loss helps to improve the model's ability to capture style texture and achieve better style transfer (DualAST).

\begin{figure*}
\centering
\includegraphics[width=0.75\textwidth]{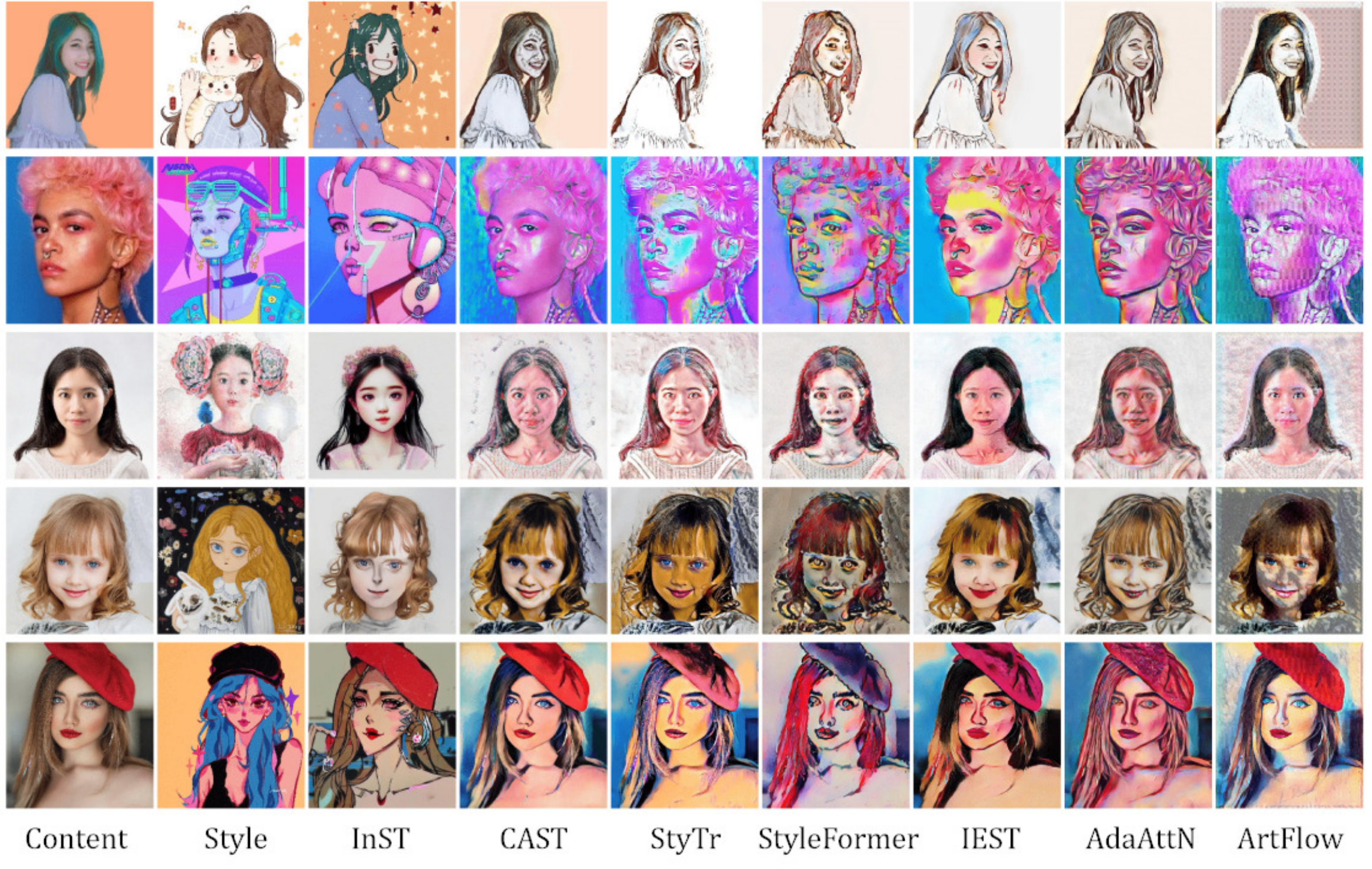} % Reduce the figure size so that it is slightly narrower than the column.
\caption{Style transfer with different network structures. InST \cite{Zhang2023inst} is implemented based on Diffusion model. CAST \cite{zhang2022domain}, StyTr \cite{deng2022stytr2}, StyleFormer \cite{wu2021styleformer}, IEST \cite{chen2021artistic}, AdaAttN \cite{liu2021adaattn} and ArtFlow \cite{an2021artflow} are implemented based on convolutional neural networks or transformer. The generated effect comes from InST \cite{Zhang2023inst}.}
\label{fig6}
\end{figure*}

\textbf{Effect evaluation.} The authenticity of the images generated by the image translation model is mainly evaluated in a domain manner. All images are fitted into a distribution, and the generation effect is measured by the similarity between the generated images and the real images fitting distribution. But in image translation, the preservation of the generated image structure is evaluated in the form of a single image. Compared with image translation, style transfer involves more types of images. The wide variety of images is difficult to fit with a good data distribution. The evaluation is concentrated on a single image, and the final result is often the average of the results of multiple images.

In summary, the application scope of image translation is small and detailed, and it can achieve conversion with large semantic differences. The application scope of style transfer is wide, but it focuses more on the fusion of texture features. In order to expand the conversion capability of image translation and try to break through the limitations of the content structure in the domain, some models have begun to appear to deal with this aspect. Generally speaking, the stylization of scenes and faces is achieved through different model structures due to their huge content structure differences. GMU \cite{men2022unpaired} uses refined Gated Cycle Mapping to achieve the stylization of scenes and faces through a single model. This method has greatly promoted the application scope of image translation. Recently, the application of diffusion models \cite{ho2020denoising} has also gradually narrowed the gap between the two technologies. For example, InST \cite{Zhang2023inst} is similar to style transfer technology, but it can achieve realistic image shape changes, which is not easily achieved by architectures such as convolutional neural network and Transformer. As shown in Figure \ref{fig6}, only InST achieves exaggerated shape changes, while the other models based on convolutional neural network and Transformer architectures are more about texture transfer. This shows the huge potential for style transfer in the future. Style transfer becomes more powerful based on its wide application by combining with the Diffusion model, but the style images processed by the single InST model are limited. Starting from 2022/2023, image generation technology has developed extremely rapidly \cite{rombach2022high,kumari2023multi}, and more and more people are paying attention to the generation effects of multiple scenes/general categories. This requires image translation and style transfer to make a lot of efforts to keep up with technological progress. In the future, the evolution of technology will show more application models of arbitrary image processing, including texture and color transfer, as well as semantic information transfer.

\section{Conclusion}

In this paper, we discuss the main differences and connections between image-to-image translation and style transfer from multiple aspects such as technology type, training mode and effect evaluation. Through the discussion in this paper, the boundaries between the two technologies can be clearer and better serve the community. Image translation mainly focuses on the conversion between images with similar content structures, which can achieve larger shape changes. Style transfer covers a wider range and can include any content image and style image, but most of them achieve changes in texture and color distribution, and cannot achieve larger shape changes. With the development of generative technology, models combined with Diffusion model have emerged, and style transfer has begun to be able to transform the shape of images. Due to the rapid development of general technology, higher requirements are placed on the breadth of technology application. Focusing on texture changes and semantic information changes in a wider range of scenarios has greater development prospects in the future.

\bibliography{aaai25}% common bib file

\end{document}